%% file: main.tex
\documentclass[letterpaper]{article}

\usepackage{natbib,alifeconf}  

\usepackage{amssymb}
\usepackage{url}
\usepackage{mathtools}
\usepackage{graphicx}
\usepackage{ragged2e}
\usepackage{tabularx}
\usepackage{hyperref}
\title{Decentralized scheduling through an adaptive, trading-based multi-agent system}
\author{Michael Kölle$^{1}$, Lennart Rietdorf$^{1}$ \and Kyrill Schmid$^{1}$ \\
\mbox{}\\
$^1$LMU Munich, Germany \\
\{michael.koelle, kyrill.schmid\}@ifi.lmu.de, len.rietdorf@campus.lmu.de} 

\begin{document}
\maketitle

\input{content/0-abstract}
\input{content/1-introduction.tex}
\input{content/3-related-work.tex}
\input{content/4-approach.tex}
\input{content/5-evaluation.tex}
\input{content/6-conclusion.tex}

\footnotesize
\bibliographystyle{apalike}
\bibliography{main} 

\end{document}

%% file: content/0-abstract.tex
\begin{abstract}
In multi-agent reinforcement learning systems, the actions of one agent can have a negative impact on the rewards of other agents. One way to combat this problem is to let agents trade their rewards amongst each other. Motivated by this, this work applies a trading approach to a simulated scheduling environment, where the agents are responsible for the assignment of incoming jobs to compute cores. In this environment, reinforcement learning agents learn to trade successfully. The agents can trade the usage right of computational cores to process high-priority, high-reward jobs faster than low-priority, low-reward jobs. However, due to combinatorial effects, the action and observation spaces of a simple reinforcement learning agent in this environment scale exponentially with key parameters of the problem size. However, the exponential scaling behavior can be transformed into a linear one if the agent is split into several independent sub-units. We further improve this distributed architecture using agent-internal parameter sharing. Moreover, it can be extended to set the exchange prices autonomously. We show that in our scheduling environment, the advantages of a distributed agent architecture clearly outweigh more aggregated approaches. We demonstrate that the distributed agent architecture becomes even more performant using agent-internal parameter sharing. Finally, we investigate how two different reward functions affect autonomous pricing and the corresponding scheduling.
\end{abstract}

%% file: content/1-introduction.tex
\section{Introduction}
\label{sec:introduction}
In the field of cooperative AI, we seek methods to establish cooperative behavior amongst independent and autonomous agents \citep{dafoe2020open}. Many situations, like autonomous driving, require autonomous agents working together, which makes the ability to act cooperatively a key part in integrating artificial intelligence into our daily lifes.
Research in reinforcement learning (RL) has shown successful applications for single agent \citep{mnih2015,silver2017mastering} and multi-agent systems \citep{leib17, phan2018leveraging, vinyals2019grandmaster}. While fully cooperative tasks, where all agents receive the same reward and pursue the same goal, can be solved by a centralized training approach, this is not the case if agents have independent rewards and goals. Furthermore, the behavior of purely self-interested agents in multi-agent systems with common, shared resources often results in sequential social dilemmas which expose tension between individual and collective rationality \citep{Rapoport1974}, especially when the resources are scarce \citep{leib17}.
Exploration in this field has led to various approaches on how to influence the self-interested actions of independent, decentralized training agents towards a higher, emergent common good. The approaches range from game theory \citep{lerer2017maintaining}, modeling of social preferences \citep{buso10}, to ones where agents incentivise each other to cooperate \citep{yang20, schm18, lupu20, schm21}. 

In this work, we build upon the action market approach in \cite{schm18,schm21}, specifically we introduce a step between making and accepting an offer. This allows agents to observe the offers first and then make a decision about those offers whereas in \cite{schm18} the agents had to guess the demand based on past transitions. Because of this extra step, the emergent cooperation tends to be more stable than in \cite{schm18}. 
 We introduce a multi-agent scheduling environment where mutual benefits can be realized by trading with each other. In this environment highly multi-dimensional actions are necessary to trade and allocate jobs on compute cores. Using neural networks (NNs) as the decision making entities, an important problem emerges: Each multi-dimensional action has to be translated into one of exponentially many one-dimensional actions. This renders decision making exponentially difficult with a linearly increasing number of compute cores and queue lengths. Therefore, we evaluate different agent architectures that are designed to address or even circumvent the problem of exponential action spaces. Thus, we answer the question which agent architecture is most successful in mastering the highly multi-dimensional action space of the trading-based scheduling environment. Additionally, we evaluate the implications if one of these agent architectures is enabled to set trading prices freely on its own. All code and parameters of the experiments can be found here\footnote{\href{https://github.com/lr40/marl-scheduling.git}{https://github.com/lr40/marl-scheduling.git}}.

%% file: content/3-related-work.tex
\section{Related Work}
\label{sec:related-work}
In this work, the approach to coordinate the multiple agents is related to the Action Market introduced in \cite{schm18, schm21}. Yet there is an important difference in the time dimension: While the market mechanism in \cite{schm18} is based on a simultaneous matching of supply and demand, the market mechanism of this work always has a time step between the making of an offer and its acceptance. Agents observe first and then decide which ones to accept and which ones not to accept. The "guessing" of a demand, learned by past rewards as in \cite{schm18}, which then leads to behavioral changes, does not exist here. Rather an explicit offer, which will be observed, has to be accepted to conclude the trade. This circumvents the problem that agents in \cite{schm18} learn over time to take advantage of the cooperative behavior of the counter party: They do so by suddenly stopping the costly demand action - even though demand is still present. The deceived agent then delivers the desired action - expecting the traded reward - without receiving the reward. Learned cooperation can be unlearned by such breaches of trust. On the other hand, the approach presented in this paper is closer to a direct, explicit communication of the agents. Thus, it is accompanied by an increased overhead.

%% file: content/4-approach.tex
\section{Approach}
\label{sec:approach}
\subsection{Scheduling environment}
Scheduling comes into play when the demand for resources is higher than the available processing capacity \citep{tane09}. In this context we use scheduling for the assignment of computational jobs to computational cores. A computational job $i$ can be described by its arrival time $t^{AT}_i$, burst time $t^{BT}_i$ and priority $p_i$. The burst time is the duration it takes for the job to be executed. This time cannot be influenced by the scheduler. After the jobs have completed we measure the turnaround time $t^{TAT}_i$ (Eq. \ref{eq:tat + ntat}) for each job. The turnaround time is the total duration that the job remained in the system. In addition to the burst time, it also includes the waiting time $t^{WT}_i$ during which the job waited inactively for its allocation to a core. 
\begin{equation}
\label{eq:tat + ntat}
    t^{TAT}_i = t^{BT}_i + t^{WT}_i \; \; \; \; \; \; \; \; \; \; \; \; \; \; \;  t^{NTAT}_i = \frac{t^{TAT}_i}{t^{BT}_i}
\end{equation}
In order to evaluate a scheduling method, we use the burst time and turnaround time to measure the normalized turnaround time $t^{NTAT}_i$ (Eq. \ref{eq:tat + ntat}) of a computational job. The scheduling process controls the waiting time. The closer the normalized turnaround time approaches its optimal value 1, the smaller the included waiting time has been.

\begin{figure}[ht]
  \centering
  \includegraphics[width=\linewidth]{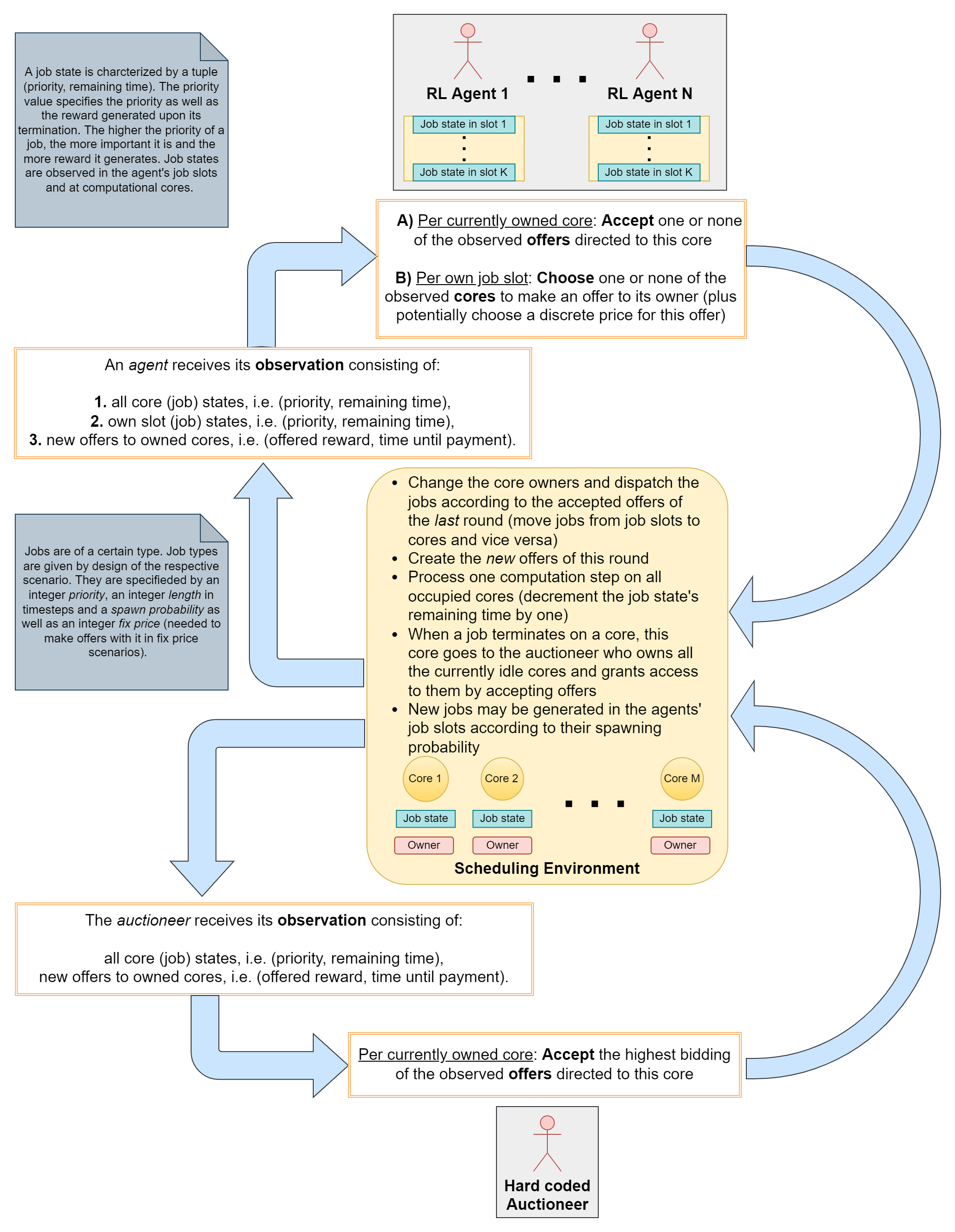}\\
  \caption[General structure of the environment]{The general structure of the RL environment. Agents own cores when they are currently computing on them. If no compute job is currently running on a core, the auctioneer owns it. Agents make offers to the owner of a core to let a job compute on that core if the offer is accepted. }
  \label{fig: environment loop}
\end{figure}

We implemented a scheduling environment for multiple RL agents. In this environment a common resource - \textit{M} computing cores - ought to be used efficiently by \textit{N} agents, each with \textit{K} job slots. The agents want to compute renewing jobs from the slots on the cores. Jobs are of a certain type whose combination of priority value, length in timesteps and spawn probability are specified as an environment parameter. The trading mechanism allows the agents to enhance the individual rewards as well as the overall scheduling performance. In contrast to traditional, centralized scheduling concepts, the scheduling of this work is partially decentralized. The agents have the chance to exchange access to compute cores amongst each other if the current owner (i.e. user) of a core accepts an explicit, observed offer directed at this core. This offer is a 2-tuple consisting of the reward payment and the necessary timesteps until payment. In scenarios with fixed prices, the reward payments are given as an environment parameter corresponding to the job type's priority. In scenarios with free prices, the reward payment can be chosen by the offering agent.
No scheduling strategy is predefined; the scheduling results from the learned actions of the agents. An exception to this is the hard coded auctioneer, which manages currently idle cores (agents lose ownership of idle cores by design) and grants access to the highest bidding RL agent. The pure behavior of the auctioneer - without any intra-agent trading - implements a variant of first-come-first-serve (FCFS) where jobs with higher priority values are preferred. Figure \ref{fig: environment loop} gives an overview of the scheduling environment and the RL loop that the environment implements.

\subsection{Rewards}
\label{sec:rewards}
In the scheduling environment, the rewarding scheme for the individual agents follows an egoistic principle. This is plausible in the context of independent learners that want to get their jobs computed or want to receive traded rewards.
\subsubsection{Reward for terminating jobs and trading access}
\label{subsec: acceptor rewards}
The responsible owner of a core - i.e. the agent currently running a job on it - gets a reward as soon as its own job has terminated. This reward has the same value as the priority of the job. The owner of a core can receive a reward in yet another way: through a trade. If offers are accepted, the promised reward payments are first stored by the environment in a chronological order with respect to a core as long as no job has terminated yet. When a job terminates all stored reward claims and liabilities of the preceding trades are settled and each participant in the reward chain receives its net payout. These reward chains initially start with the auctioneer as the owner of all cores and bring no additional net rewards into the system when a job terminates. They are only a redistribution of the generated reward.
\subsubsection{Reward for making offers}
\label{subsec: core chooser rewards}
For offering, on the other hand, the rewards are received directly in the next time step if the offer has been successfully accepted by the counterparty. The generated reward is equal to the priority of the mediated job.
\subsubsection{Price setter reward}
\label{subsec: price setter rewards}
In free price scenarios, the agents will be enabled to freely set prices for their offers. The rewarding scheme of the price setting action can be motivated by a commercial and a non-commercial motive. In the former case the agent wants to bid an amount as small as possible that is just on the brink of being accepted. Mainly, the paid out reward is the difference between the priority of the facilitated job $p_i$ and the chosen price $x$ which can also be negative if the price is set too high (Eq. \ref{eq:reward_func_1}).

\begin{equation}
\label{eq:reward_func_1}
   R_{1}(p_i, x) = \begin{dcases*}
                p_i - x & if $p_i \neq x$\\
                0.5 & otherwise
\end{dcases*} 
\end{equation}

In the non-commercial case, the reward equals the priority of the the facilitated job $p_i$ as long as the set price $x$ was not higher than the job priority (Eq. \ref{eq:reward_func_2}). In this case the price setter can increase the bids for free as long as it does not overshoot.

\begin{equation}
\label{eq:reward_func_2}
   R_{2}(p_i, x) = \begin{dcases*}
                p_i & if $p_i \geq x$\\
                p_i - x & otherwise
\end{dcases*} 
\end{equation}

\subsection{Agent architectures}
\label{sec:agents structures}
Multi-agent systems consist of multiple agents that share a common environment. An agent is an autonomous entity with two main capabilities: observing and acting. The observation of the current state of the environment allows the agent to choose an appropriate action out of a given action set. The chosen action depends on an agent's policy. In this work, we used different RL agent architectures based on the PPO algorithm \citep{SchulmanWDRK17} to learn a good policy for the proposed environment. All NNs used in this work contain one hidden layer of neurons. An important constraint to consider in this regard is that RL algorithms usually allow one NN to output only one one-dimensional (1D) action at a time. A multi-dimensional action consisting of $d$ sub actions is only attainable if the environment translates the chosen 1D-action $a$ back into one of the $\mathcal{O}(2^{d})$ $d$-dimensional actions. In this manner, the resulting amount of actions grows exponentially as the dimensionality of the multi-dimensional action increases with the problem size or scaled setup of the environment.
\subsubsection{Fully aggregated agent}
Disregarding this important constraint, one can naively construct an agent composed of only one neural net that has one hidden layer of 64 neurons. The action of this single neural net is responsible for accepting or declining offers for up to $M$ cores (if this agent owns all the cores), and making up to $K$ offers for all of its associated $K$ job slots at the same time. A single offer is made by choosing a target core for a job of an own slot. The prices are predefined per job type as an environment parameter. The dimensionality of this aggregated action vector sums up to overwhelming $M+K$ which translates to $\mathcal{O}(2^{M+K})$ many actions. The agent type implementing this design will henceforth be called \textit{fully aggregated agent}.
\subsubsection{Semi-aggregated agent}
The first step to reduce the exponential scaling behavior of the action space is to split the agent up in two neural networks (each with a hidden layer of 32 neurons) which can be regarded as two independent, complementary sub agents: one part \textbf{A} for accepting resp. declining offers (cf. fig. \ref{fig: environment loop} action part \textbf{A}) and one part \textbf{B} for making offers (cf. fig. \ref{fig: environment loop} action part \textbf{B}). Core (job) states plus an ownership flag and offers to this core are observed by part \textbf{A}. Core (job) states and slot (job) states are observed by part \textbf{B}. The two networks generate two 1D-actions which will be translated back into an $M$-dimensional action (one dimension for each core) resp. $K$-dimensional action (one dimension for each job slot). This results in a less intense scaling behavior of the action space which is nevertheless still exponential. The constructed agent type will be called \textit{semi-aggregated agent}.
\subsubsection{Distributed agent}
The idea to split the agent up in several, independent, complementary neural networks can be taken one step further by dividing the semi-aggregated accepting side \textbf{A} up into $M$-many acceptor networks and dividing the offer side \textbf{B} up into $K$-many offer networks (in both cases each with a hidden layer of 16 neurons). Each acceptor network is responsible for managing the incoming offers of \textit{one} core if the agent owns the core. Each core chooser network is responsible for making the offers of \textit{one} job slot. Although the amount of the neural networks increases sharply by this, one big advantage arises: The $M+K$-many 1D-actions of the $M+K$ networks do not need to be translated into a multi-dimensional action vector. This is because these actions are already intrinsically one-dimensional since just one offer index respectively one core index has to be selected per action.
\subsubsection{Distributed agent with parameter sharing}
For the distributed agent, a possible optimization is local parameter sharing. Instead of sustaining $M+K$ independent networks which constitute the distributed agent, we implement a local parameter sharing between the $M$-many accepting networks and between the $K$-many offering networks. Consequently, only two neural networks that output in total $M+K$ intrinsic 1D-actions will make up the agent. By doing so, a faster training process and less overhead arises. Thus, the observation and action spaces correspond to atomic actions rather than aggregated actions which had to be translated to the atomic level.

\subsubsection{Distributed agent with free price setting}
So far, the considered agent types are able to make offers and to accept offers, but they are not able to set the prices for the offers on their own. Instead, they have to rely on fixed prices specified as an environment parameter. To be able to set prices freely the distributed Agent is extended by a third type of neural network: a price setter network. The chosen price has to be an integer from [0,\textit{max\_Prio}] where \textit{max\_Prio} denotes the maximum priority of all job types in the run scenario.

%% file: content/5-evaluation.tex
\section{Experiments\protect\footnote{The used hyperparameters and environment parameters can be found in the README file of the \href{https://github.com/lr40/marl-scheduling.git}{repository}.}}
\label{sec:evaluation}
\subsection{The effect of intra-agent trading}
\label{subsec:intra-agent trading}
\begin{figure}
\includegraphics[width=\linewidth]{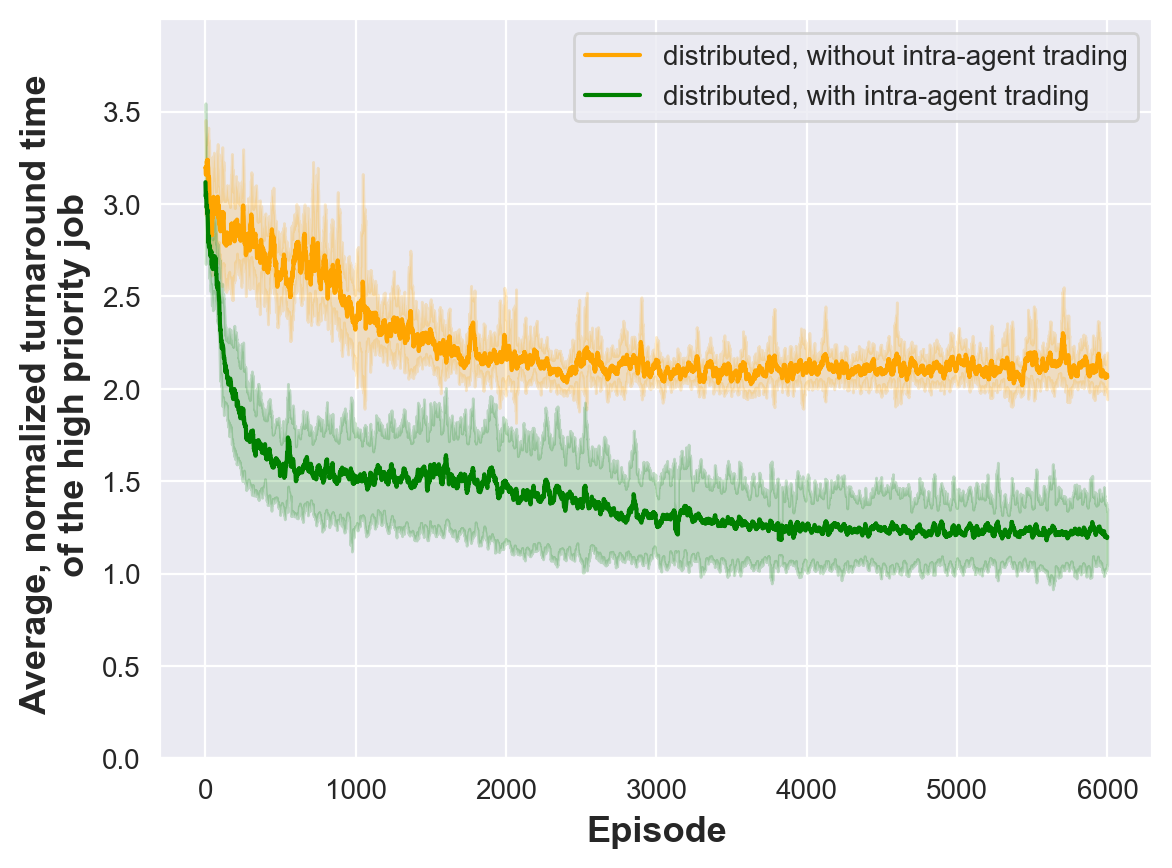}
\caption{Mean and standard deviation of 10 independent runs. Intra-agent trading significantly reduces the normalized turnaround time of the highly prioritized job type. The distributed agent architecture was used.} 
\label{fig: Experiment 1 no trading}
\end{figure}

We evaluate two types of jobs in this scenario: Long-running, frequently occurring, low-priority jobs that block the scarce compute cores and much rarer, shorter, high-priority jobs. The exchange prices are given as an environment parameter according to the job's priority. The intra-agent trading mechanism enables the agents to trade computing core access amongst each other. Figure \ref{fig: Experiment 1 no trading} shows that intra-agent trading significantly reduced the normalized turnaround time of the highly prioritized job type. Trades were used to let the high priority jobs get access to the cores. The intra-agent trading thus outperformed the priority oriented FCFS approach of the auctioneer.

\subsection{Agent architecture and scheduling performance}
\label{subsec: Agent architecture and scheduling performance1}
\begin{figure*}
  \begin{subfigure}{0.32\textwidth}
    \includegraphics[width=\textwidth]{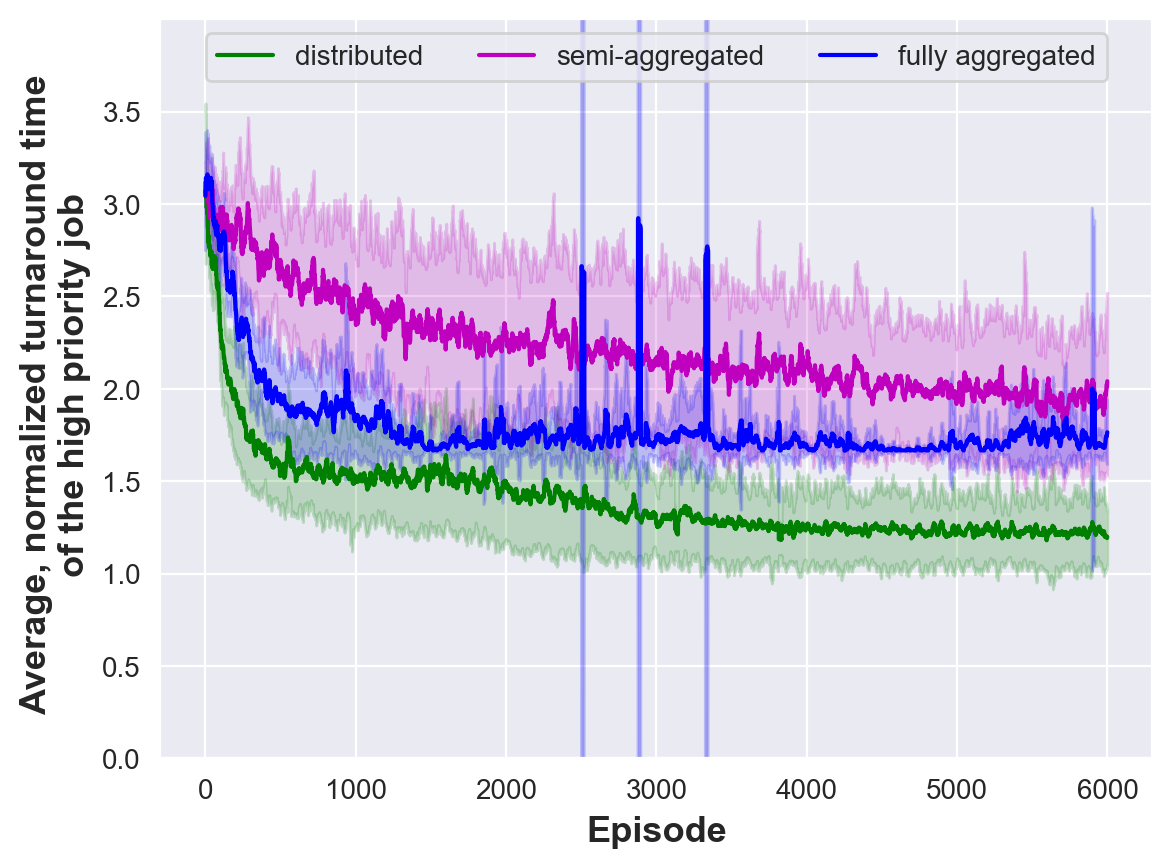}
    \caption{2 agents with self-refilling queues of 3 jobs compete for 2 cores.} 
    \label{fig: Experiment 1 2 Agenten}
  \end{subfigure}%
  \hfill
  \begin{subfigure}{0.32\textwidth}
    \includegraphics[width=\textwidth]{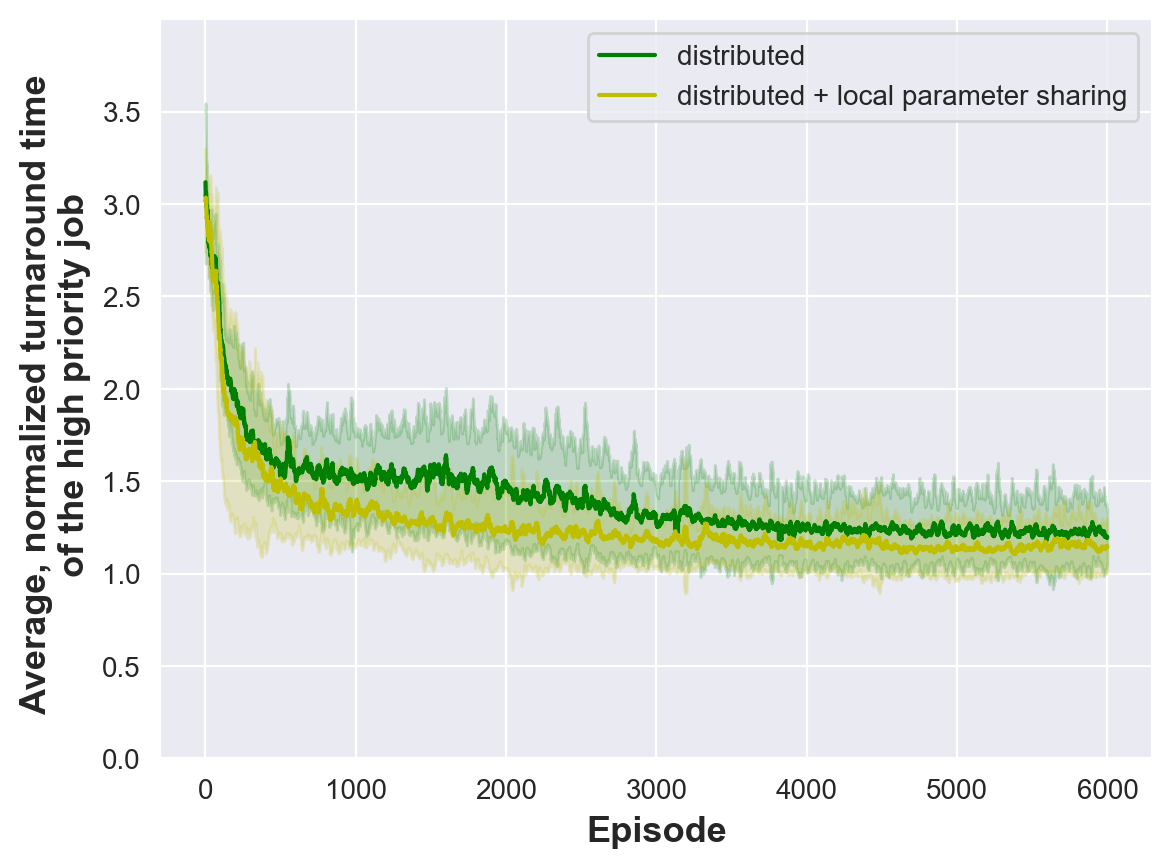}
    \caption{2 agents with self-refilling queues of 3 jobs compete for 2 cores.} 
    \label{fig: Experiment 1 2 Agenten lokales PS}
  \end{subfigure}%
  \hfill
  \begin{subfigure}{0.32\textwidth}
    \includegraphics[width=\textwidth]{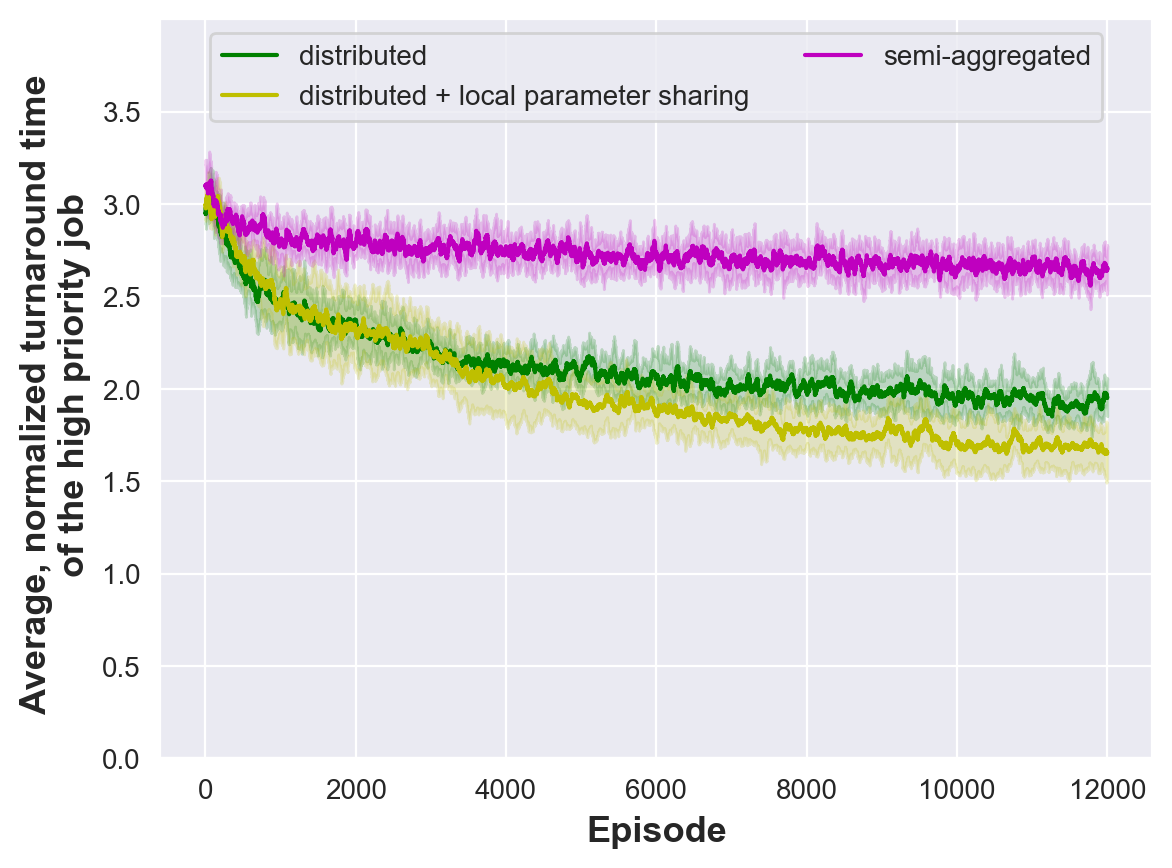}
    \caption{4 agents with self-refilling queues of 3 jobs compete for 4 cores.} 
    \label{fig: Experiment 1 4 Agenten}
  \end{subfigure}%
\caption[]{Mean and standard deviation of 10 independent runs. The average, normalized, turnaround times of the high priority job type when using the distributed (green), semi-aggregated (magenta), fully aggregated (blue) and distributed with local parameter sharing (yellow) agent architecture. The lower the curve, the better the effective scheduling performance.}
\end{figure*}

In this section the same scenario of the previous section is evaluated. Here, we used 2 agents each with 3 self-refilling job slots competing for 2 cores. Different agent architectures lead to different performances. This is due to their different sizes of action spaces. It is shown in figure \ref{fig: Experiment 1 2 Agenten} that the distributed agent architecture yields the lowest turnaround time of the high priority job type. Interestingly, the fully aggregated architecture yields the second lowest, normalized turnaround time although it has the largest action space. The better performance is achieved because the neural network learns to trade with itself since it makes the decision for accepting as well as making offers. Since the used scaling is still manageable it can derive an advantage compared to the semi-aggregated agent type (cf. table \ref{tab: cardinality of action spaces}).

Figure \ref{fig: Experiment 1 2 Agenten lokales PS} shows the same data for the distributed agent type as in figure \ref{fig: Experiment 1 2 Agenten} but compares those to the performance of the distributed agent type with local parameter sharing. It can be seen that the employment of local parameter sharing improves the overall performance as well as the time necessary for the adaption.

The scenario can also be scaled up to 4 agents each with 3 job slots who compete for 4 cores. Remarkably, this scaling is already some orders of magnitude too high for the fully aggregated agent architecture since its action space cardinality is already out of scope (cf. table \ref{tab: cardinality of action spaces}). So only the performance of the distributed architecture with and without parameter sharing as well as the semi-aggregated architecture can be compared. Figure \ref{fig: Experiment 1 4 Agenten} shows this. Again, the distributed agent architecture significantly outperforms the semi-aggregated architecture. Employing local parameter sharing yields another performance boost in addition to that.

\renewcommand{\arraystretch}{1.3}
\renewcommand\tabularxcolumn[1]{>{\Centering}m{#1}}
\begin{table}[htpb]
\small
\centering
\begin{tabularx}{\linewidth}{lXX}
\hline
\textbf{Acting unit} & \textbf{2 agents, 2 cores} & \textbf{4 agents, 4 cores} \\
\hline
distributed offer unit & $3$ & $5$ \\
distributed acceptor unit & $7$ & $13$ \\
semi-aggregated offer unit & $9$ & $125$ \\
semi-aggregated acceptor unit & $49$ & $28\ 561$ \\
fully aggregated unit & $441$ & $35\ 701\ 125$ \\
\hline
\end{tabularx}
\caption[Experiment 1: Cardinality of the action spaces of the actung (sub) units per scaling of the scenario]{Cardinality of the action spaces of the acting (sub) units for two scalings of the scenario.}
\label{tab: cardinality of action spaces}
\end{table}

\subsection{Price level and scarcity}
\label{sec:price-level-and-scarcity}
In real life, prices are expressions of relative scarcities. This experiment investigates to what extent in the scheduling environment scarcer computational cores lead to higher prices if free price setting is enabled. For the price setters, two reward regimes are compared. To keep complexity low, there is only a single job type in this scenario. Its priority and length equals 5. 2 agents each with 3 job slots - i.e. a total of 6 jobs - compete for 2 cores in one scenario and for 4 cores in the other scenario.
\begin{figure}
  \begin{subfigure}{0.52\linewidth}
    \includegraphics[width=1\linewidth]{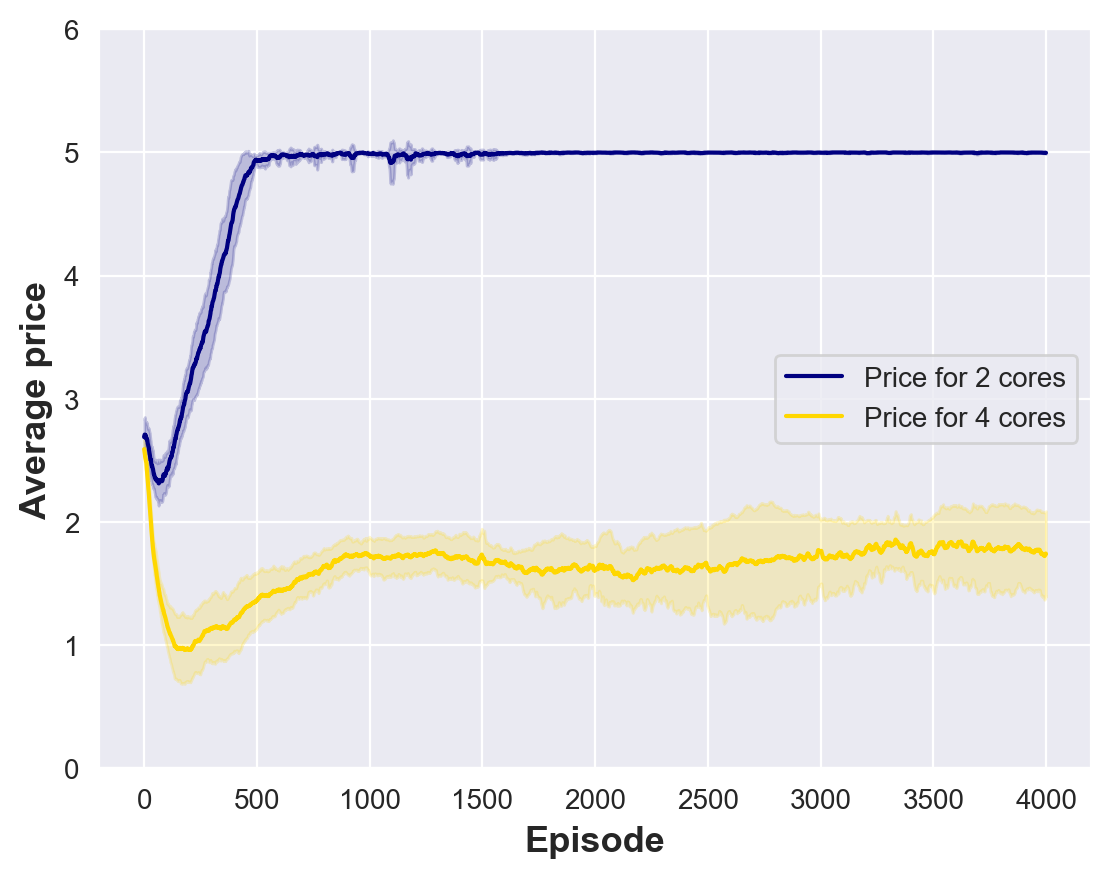}
    \caption{The price development \\ under \textbf{commercial} reward\\ when the cores are scarce (blue)\\ or less scarce (gold).} \label{fig: Experiment Scarcity a}
  \end{subfigure}%
  \begin{subfigure}{0.52\linewidth}
\includegraphics[width=1\linewidth]{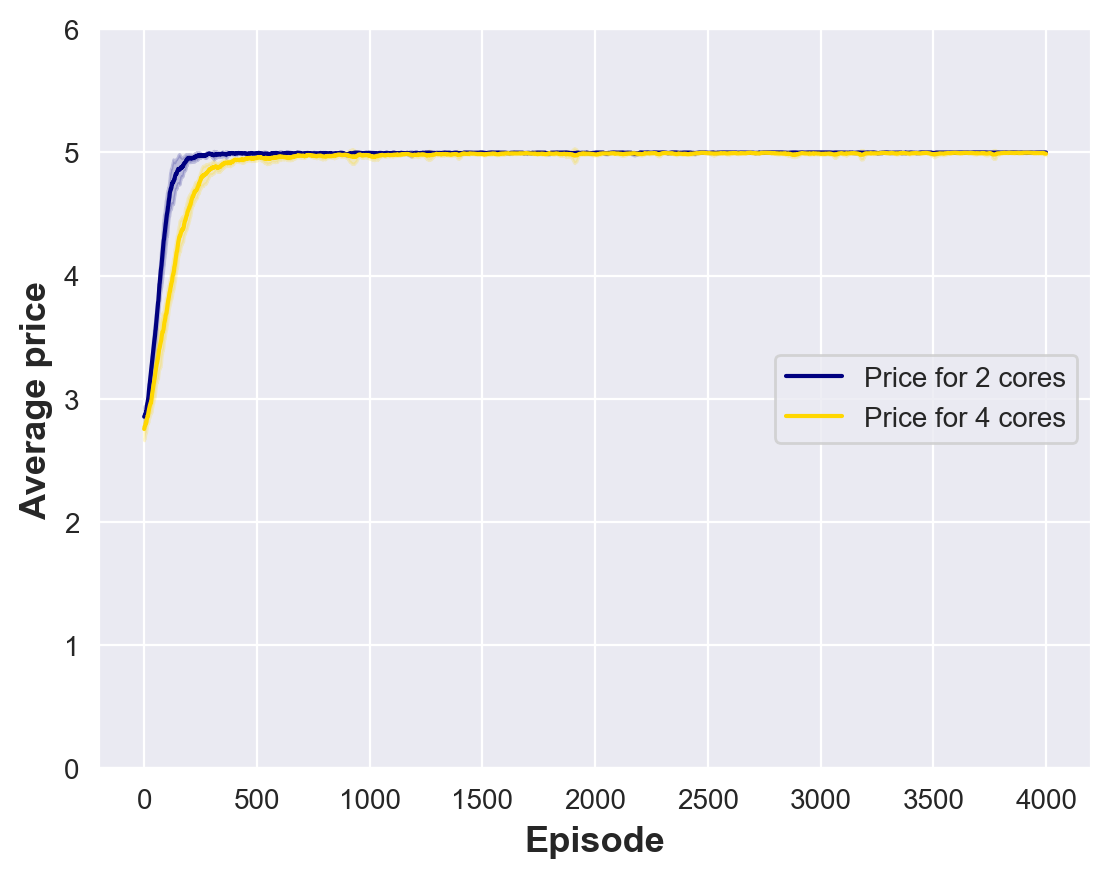}
    \caption{The price development under \textbf{non-commercial} reward when the cores are scarce (blue) or less scarce (gold).} 
    \label{fig: Experiment Scarcity b}
  \end{subfigure}%
\caption[]{Mean and standard deviation of 10 independent runs are shown.} \label{fig: Experiment Scarcity}
\end{figure}

In Figure \ref{fig: Experiment Scarcity a}, the price development in the commercial reward regime is shown with scarce 2 compute cores and more abundant 4 cores. For 2 compute cores the price rises to the maximum priority of the scenario. The 6 price setters learn to bid each other up to this level because access to the few computational cores is so contested. On the other hand, when the number of cores is doubled to 4, the commercial price setters learn a significantly lower price level. So using the commercial reward function, the found price level increases with the scarcity of computational cores.

Figure \ref{fig: Experiment Scarcity b}, on the other hand, shows the price development in the non-commercial reward regime in the same scenario with 2 and 4 cores. In both cases, the price level rises to the maximum value. In contrast to the commercial reward regime, the price also becomes maximum when more abundant 4 cores are available. This is due to the fact that price increases are not associated with any cost for the price setter but increase the chances for an offer to be accepted. Thus the high price becomes the dominant strategy. Using the non-commercial reward function, prices reflect the priority of the associated jobs rather than the scarcity of the compute cores.

\subsection{Price level and scheduling}
\label{subsec: price level and scheduling}
How the free price setting influences the realized scheduling will be analysed in this section. There are 2 agents, each with a queue length of 3 jobs that want to compute three, equally frequent job types with the priorities 2, 4, and 8 on two cores.

\begin{figure}
  \begin{subfigure}{0.52\linewidth}
    \includegraphics[width=\linewidth]{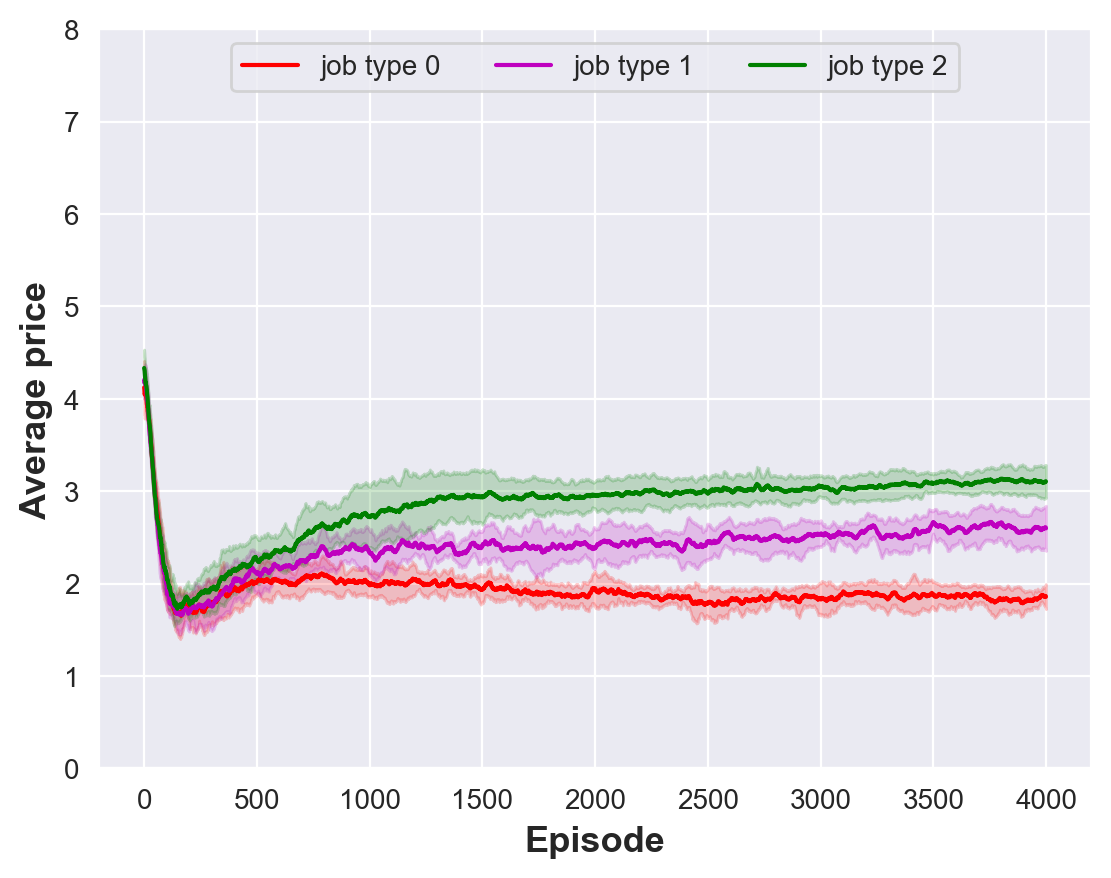}
    \caption{\textbf{Commercial} reward:\\ average prices} \label{fig: Experiment price scheduling a}
  \end{subfigure}%
  \begin{subfigure}{0.52\linewidth}
    \includegraphics[width=\linewidth]{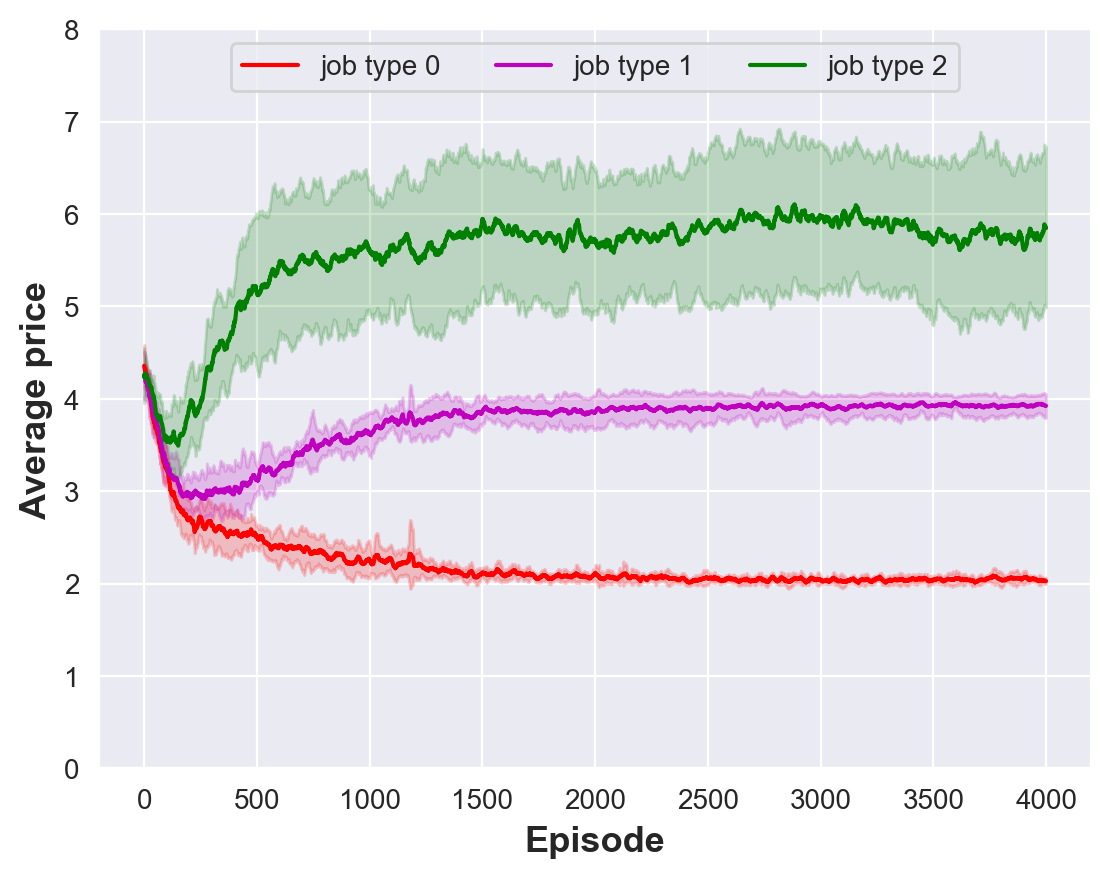}
    \caption{\textbf{Non-commercial} reward:\\ average prices} \label{fig: Experiment price scheduling b}
  \end{subfigure}%
  \newline
  \begin{subfigure}{0.52\linewidth}
    \includegraphics[width=\linewidth]{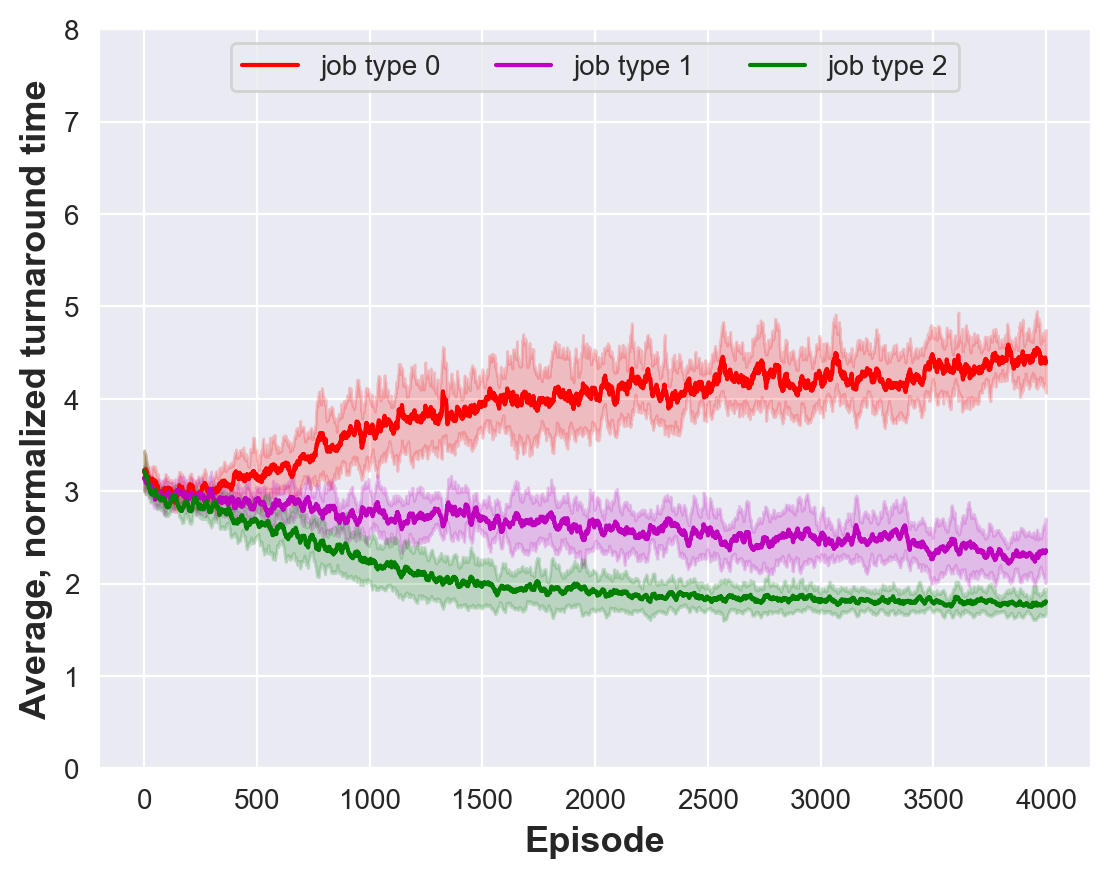}
    \caption{\textbf{Commercial} reward:\\ average, normalized\\ turnaround times} \label{fig: Experiment price scheduling c}
  \end{subfigure}%
  \begin{subfigure}{0.52\linewidth}
    \includegraphics[width=\linewidth]{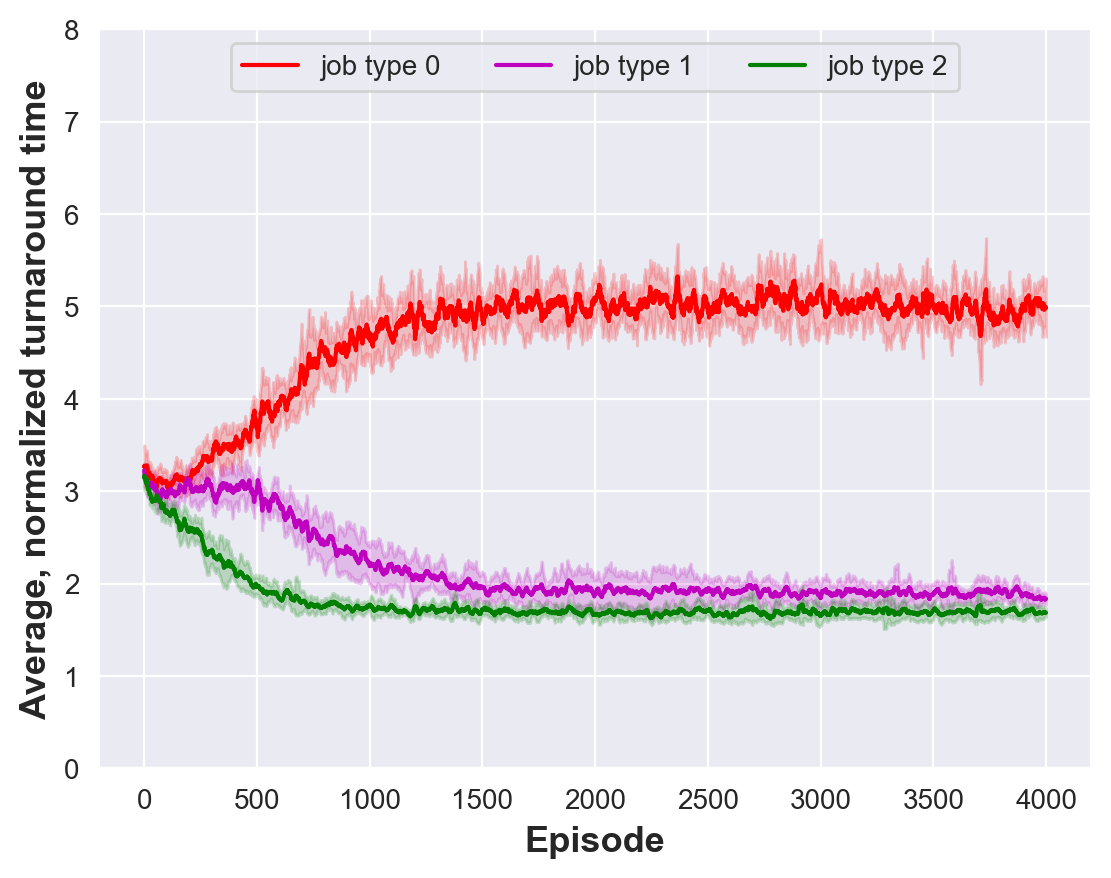}
    \caption{\textbf{Non-commercial} reward:\\ average, normalized\\ turnaround times} \label{fig: Experiment price scheduling d}
  \end{subfigure}%
  \caption[Prices and turnaround times under (non-)commercial reward]{Mean and standard deviation of 10 independent runs are shown. The priorities of the job types are given as [2 (red), 4 (magenta), 8 (green)]. The subfigures \textbf{(a)} and  \textbf{(b)} only include realized prices of accepted offers.}
\end{figure}

How prices evolve under the commercial reward regime is shown in Fig. \ref{fig: Experiment price scheduling a}. In the beginning, there is a sharp initial drop in the price levels of all job types. At a certain point, however, the commercial price setters learn that outbidding competitors is a decisive factor for success. As a result, prices for all types of jobs rise again but split up according to their priority. The transitive order of prices then corresponds to the transitive order of job type priorities. Overall, the spectrum of prices spreads less upward than for the non-commercial rewards (see subfig. \ref{fig: Experiment price scheduling b}). This is because in the case of commercial rewards there is an incentive for the individual price setter to outbid the lower job types with an amount as low as possible. The price of the lowest priority job type 0 is permanently very close to its priority of 2. At the lower end of the possible priorities, the prices have to be at their maximum to have any chance of competing with the same or other job types.

How prices evolve under the non-commercial reward regime is shown in Figure \ref{fig: Experiment price scheduling c}. Here, too, we see an initial drop in all price levels in the beginning. Compared to the commercial reward regime, however, the drop is less deep because there is less incentive for prices to fall. After that, prices also split up according to the respective job priorities when the price setters learn to outbid each other. However, the splitting leads to significantly higher prices for job types 1 and 2. This is because the price setters under this reward function get the same reward for all prices that are not too high. Hence a strong incentive for price raises is given. In the emerging system the low and medium prioritized jobs types 0 and 1 have to be priced at their maximum in order to have a chance against the arbitrarily high prices of job type 2.

The evolution of prices is reflected in the evolution of turnaround times, shown in Figures \ref{fig: Experiment price scheduling c} and \ref{fig: Experiment price scheduling d}. Under both reward functions, as the prices of higher-priority jobs increase, the corresponding turnaround times decrease and vice versa. The greater spread of price levels under non-commercial rewards causes the turnaround times to spread apart more sharply and quickly. For commercial rewards, on the other hand, the normalized turnaround times are closer together and change only more slowly. This slower and weaker adaptation is caused by the less differentiated prices.

%% file: content/6-conclusion.tex
\section{Conclusion}
\label{sec:conclusion}
This work addressed the question which agent architecture is most successful in mastering the highly multi-discrete action space of the trading-based scheduling environment. Success depends on how the agents' action and observation spaces scale, which in turn is largely determined by the agent architecture. Semi- and fully aggregated agent types exhibit an exponential scaling behavior of their observation and action spaces with the scenario size. This exponential scaling behavior can be transformed into a linear one if the agent is split up in multiple neural networks. We show that agents of the distributed architecture adapt best and fastest to a scenario where high priority jobs are to be prioritized over low priority ones via intra-agent trading. Furthermore, it became evident that the action space of the fully aggregated agent type increased so strongly even at small problem sizes that it is unsuitable for practical use. In addition, we show that the performance of the distributed architecture further improved when agent-level parameter sharing was implemented.

Finally, we examined the results when agents set the trading prices freely by themselves. We compared the effects of two, different reward functions for the price setter. Using the commercial reward function, the emergent prices can reflect the scarcity conditions in the environment, whereas the non-commercial reward function causes prices to be a neutral carrier of information about job priorities. For scheduling different job types it was found that adaption succeeded under both reward functions. The higher the priority of a job type, the lower its mean normalized turnaround time and the higher its found price. Compared to the commercial reward, the non-commercial reward led to more differentiated prices, lower turnaround times of high priority jobs and faster adaption.

In conclusion, splitting the RL agent into different submodules led to crucial performance improvements in the trading-based scheduling environment compared to more aggregated approaches. The reduced cardinality of the action spaces even allowed for the successful introduction of free price setting by the agents.

\clearpage